\documentclass[letterpaper, 10 pt, conference]{ieeeconf} 
\IEEEoverridecommandlockouts                              
\usepackage{graphicx}      
\usepackage{algorithm} 
\usepackage{algpseudocode} 
\usepackage{varwidth} 
\usepackage{amsfonts}
\usepackage{amsmath}
\usepackage{url}
\usepackage{bbm} 

\usepackage[caption=false,font=footnotesize]{subfig}

\title{\LARGE \bf
A Learning Framework for Robust Bin Picking by Customized Grippers
} 
\author{Yongxiang Fan, Hsien-Chung Lin, Te Tang, Masayoshi Tomizuka 
\thanks{Yongxiang Fan and Masayoshi Tomizuka are with
        University of California, Berkeley
        {\tt\small {yongxiang\_fan, tomizuka}@berkeley.edu}}%
\thanks{Hsien-Chung Lin and Te Tang are with FANUC Advanced Research Lab. }
\thanks{This work was supported by FANUC Corporation.}
}

\begin{document}
\maketitle
\thispagestyle{empty}
\pagestyle{empty}

\begin{abstract}
Customized grippers have specifically designed fingers to increase the contact area with the workpieces and improve the grasp robustness. However, grasp planning for customized grippers is challenging due to the object variations, surface contacts and structural constraints of the grippers. In this paper, we propose a learning framework to plan robust grasps for customized grippers in real-time. The learning framework contains a low-level optimization-based planner to search for optimal grasps locally under object shape variations, and a high-level learning-based explorer to learn the grasp exploration based on previous grasp experience. 
The optimization-based planner uses an iterative surface fitting (ISF) to simultaneously search for optimal gripper transformation and finger displacement by minimizing the surface fitting error. The high-level learning-based explorer trains a region-based convolutional neural network (R-CNN) to propose good optimization regions, which avoids ISF getting stuck in bad local optima and improves the collision avoidance performance.
The proposed learning framework with RCNN-ISF is able to consider the structural constraints of the gripper, learn grasp exploration strategy from previous experience, and plan optimal grasps in clutter environment in real-time. The effectiveness of the algorithm is verified by experiments.
\end{abstract}

\section{Introduction}
Customized grippers have been broadly applied in industry to execute complex tasks such as assembly and packaging. Compared with general parallel grippers, customized grippers usually consist of curved fingers and have large contact surfaces to match with the geometries of the objects. Therefore, the customized grippers can generate more stable and robust grasps compared with those generated by general parallel grippers. 

However, the grasp planning for customized grippers is challenging. On one hand, it has potentially large contact areas with workpieces, thus the traditional point contact model assumption~\cite{murray1994mathematical} and the quality metrics~\cite{ferrari1992planning, li1988task} built upon the point contact model are not applicable. As a result, the performance of the grasp planning algorithms based on those metrics~\cite{mahler2016dex, hang2014hierarchical} would be downgraded. 
On the other hand, the grasp policies learned from end-to-end manner~\cite{levine2016learning, pinto2016supersizing} use general parallel grippers with millions of grasping data, and changing fingertips require re-collecting data and re-training the policies. With the tend of mass customization and requirement to change fingertips frequently, the end-to-end learning becomes less efficient. 

There are a few works that consider the surface information during grasp planning. In~\cite{ciocarlie2007dexterous}, the optimal grasp is searched by enclosing the object with more contacts to match the object surface. However, it has a heavy computation load, thus is not suitable for online implementation. The grasp synthesis of human hands with shape matching is introduced in~\cite{li2007data}. However, this approach requires offline human demonstration and exhaustive search in the database. 
In~\cite{klingbeil2011grasping,ten2018using}, the curve of the parallel gripper is matched to the surface of objects to speed up the searching process. However, it cannot be applied to the precise matching for customized grippers with complicated shapes. 

In this paper, we propose a learning framework to plan grasps for customized grippers. To consider more complicated gripper shapes and retrieve reliable and secure grasps for different objects, it is desired to match the surfaces on the gripper to the object more precisely.  
We use an iterative surface fitting (ISF) algorithm proposed in our previous work~\cite{fan2018grasp} to fit the contact surface of multiple fingers to the object surfaces while satisfying the structural constraints. A guided sampling is also introduced to avoid the local optima and encourage the exploration of the regions with high fitting scores. 
ISF achieves efficient and precise grasp planning for a single object or slight clutter environment with small searching regions. However, the searching becomes less efficient in heavy clutter environment due to the excessively sampling candidates~\cite{fan2018grasp}. 

To improve the efficiency of grasp planning in a heavy clutter environment, we propose a topological learning approach using regions with convolutional neural networks (R-CNN)~\cite{girshick2014rich} to learn the essential features that affect the successful execution of grasps. 
The features include the spatial relationships between objects which are generally difficult to model. Thus the input to the CNN is the patches of the candidate regions. This learning-based planner is connected hierarchically with ISF in order to reduce the effect of object variations and plan reliable/precise grasps under data shortage.  
Therefore, the learning framework includes an optimization-based planner, which searches for precise grasp pose based on the fine details of the selected region using ISF, and a learning-based explorer, which learns the desired regions to start ISF searching. 

Compared with end-to-end learning methods~\cite{mahler2016dex, levine2016learning}, the proposed learning-based explorer ignores the details of grasp planning by detecting the desired regions within the image plane for potentially high fitting scores and better collision avoidance performance, 
thus the dimension of the learning module is lower than end-to-end learning. The optimization-based planner searches for optimal grasps precisely in the chosen region based on the object-specific features, which are generally not shared across objects and difficult to learn by end-to-end manner. 
Therefore, the proposed learning framework is able to improve the learning efficiency and performance at the same time. 
 
The contributions of this paper are as follows. First, the proposed low-level optimization-based planner includes both the fingertip surfaces and the structural constraints of the gripper such as the jaw width and the allowable degree-of-freedoms (DOFs). It also achieves simultaneous surface fitting and gripper kinematic planning. 
Second, by combining with the dedicated gripper design, the proposed surface fitting algorithm can deal with objects with complicated shapes as well as a clutter task with unsegmented point clouds. 
Furthermore, the proposed grasp exploration strategy avoids getting stuck in the local optima by learning from the previous grasping experience. The grasp planning by ISF and R-CNN achieves an efficient planning and the time to search for a collision-free grasp is 1.5 secs for the objects in heavy clutter environment. The experimental videos are available at~\cite{website}.

The remainder of this paper is described as follows. The grasp planning problem for customized gripper is stated in Section~\ref{sec:problem_statement}, followed by the introduction of the proposed learning framework in Section~\ref{sec:learning_framework}. The experimental verification of the proposed framework on grasp planning with customized grippers are presented in Section~\ref{sec:experimental_results}. 
Section~\ref{sec:conclusions} concludes the paper and proposes future works. The experimental videos are available at~\cite{website}.

\section{Problem Statement}
\label{sec:problem_statement}
\begin{figure}[t]
	\begin{center}
	    {\includegraphics[width =0.8\linewidth]{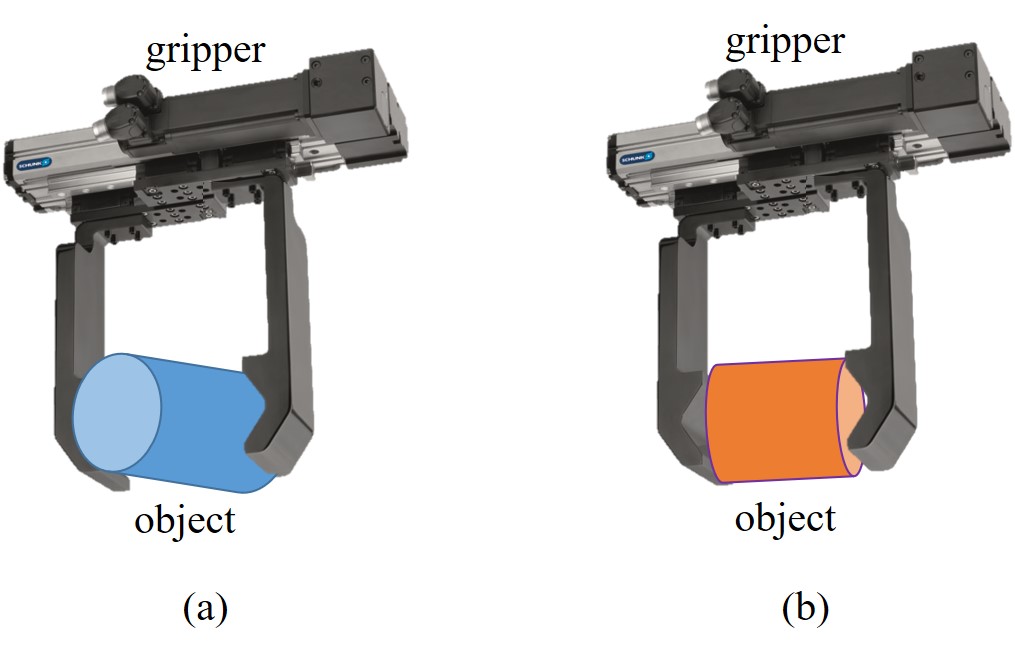}}
		\caption{An grasp example by a customized gripper with curved fingertip surfaces. A natural quality metric is the surface fitting error between the gripper and the object, where the blue object in (a) has more steady grasp compared with the orange object in (b).}
		\label{fig:gripper_object}
	\end{center}
\end{figure}

The objective of the grasp planning is to search for the optimal gripper pose and finger configuration by maximizing a quality metric considering the constraints of the customized grippers. A grasp planning example is illustrated in Fig.~\ref{fig:gripper_object}. The customized gripper has parallel jaws with customized curved fingertip surfaces. Taking the gripper in Fig.~\ref{fig:gripper_object} as an example, the grasp planning problem is defined as 
\begin{subequations}
	\label{eq:general_form}
	\begin{align}
	\max_{R, t, \delta d, \mathcal{S}_j^f, \mathcal{S}_j^o} &\  Q(\mathcal{S}_1^f, \mathcal{S}_2^f,\mathcal{S}_1^o, \mathcal{S}_2^o) \label{eq1:cost}\\
	s.t. \quad 
	& \mathcal{S}_j^f \subset \mathcal{T}(\partial \mathcal{F}_j;R,t,\delta d), \quad j = 1,2 \label{eq1:surface_finger}\\
	& \mathcal{S}_j^o = NN_{\partial \mathcal{O}} (\mathcal{S}_j^f), \quad j = 1,2 \label{eq1:surface_object}\\
	& (\mathcal{S}_1^f, \mathcal{S}_2^f) \in \mathcal{W}(d_0 + \delta d) \label{eq1:constraint1}\\
	& d_0 + \delta d \in [d_{\text{min}}, d_{\text{max}}] \label{eq1:constraint2}
	\end{align}
\end{subequations}
where $j \in \{1,2\}$ is the finger index,  $R \in SO(3), t\in \mathbb{R}^3$ are the rotation and translation of the gripper jaw, $\delta d\in \mathbb{R}$ is the jaw displacement from the original width $d_0$, and $Q$ is the quality metric in terms of the finger contact surfaces $\mathcal{S}_j^f$ and the object contact surface $\mathcal{S}_j^o$. The finger contact surface $\mathcal{S}_j^f$ lies on the finger surface $\partial \mathcal{F}_j$ transformed by $\mathcal{T}$ with amount of $R,t$, as shown in~(\ref{eq1:surface_finger}). The object contact surface $\mathcal{S}_j^o$ is defined by searching the nearest neighbor of the $\mathcal{S}_j^f$ on the object surface $\partial \mathcal{O}$, as shown in~(\ref{eq1:surface_object}).
Constraint~(\ref{eq1:constraint1}) indicates that the finger contact surfaces should be constrained in the space $\mathcal{W}$ parameterized by the jaw width, and (\ref{eq1:constraint2}) describes the finger displacement constraint.

Problem~(\ref{eq:general_form}) would be a standard grasp planning if the contact surface degenerates into a single contact point. In general, however, the point contact model may not be able to directly include gripper surface into the planning. Problem~(\ref{eq:general_form}) is challenging to solve by either learning or optimization. On one hand, the learning requires training on objects with large variety. On the other hand, the optimization with gradient or sampling can either be nontrivial to use contact surfaces as decision variables, or require to search the whole state space, which is not appropriate for real-time implementation. 

A natural surface-related quality can be constructed by matching the surfaces of the gripper towards the object, as shown in Fig.~\ref{fig:gripper_object}. Intuitively, the grasp with small surface fitting error~(Fig.~\ref{fig:gripper_object}(a)) is more reliable and robust compared with the one with large surface fitting error~(Fig.~\ref{fig:gripper_object}(b)). With the surface fitting error as the quality metric, the problem can be addressed by point set registration algorithm, such as the rigid registration (e.g. ICP~\cite{besl1992method}) or non-rigid registration (e.g. CPD~\cite{myronenko2010point} or TPS~\cite{bookstein1989principal}). The non-rigid registration allows arbitrary deformation of the point set without considering the structural constraints (Constraint~\ref{eq1:constraint1}~\ref{eq1:constraint2}), while the rigid registration assumes rigid transformation of the point set without considering the allowable motion between different portions of the points. 
Both the methods tend to be trapped into local optima during the searching. 

\section{The Learning Framework}
\label{sec:learning_framework}
\subsection{Overview}
In this paper, we assume that the customized gripper has rigid fingertip surfaces and allows motion in certain DOFs. Therefore, the proposed searching algorithm is modified from the rigid registration by including the structural constraints of different fingers such as the width and DOFs. To avoid being trapped into local optima and achieve better collision avoidance performance, we propose a learning-based explorer to detect the desired regions to initialize the low-level search. 

\begin{figure}[t]
	\begin{center}
		{\includegraphics[width=3.4in]{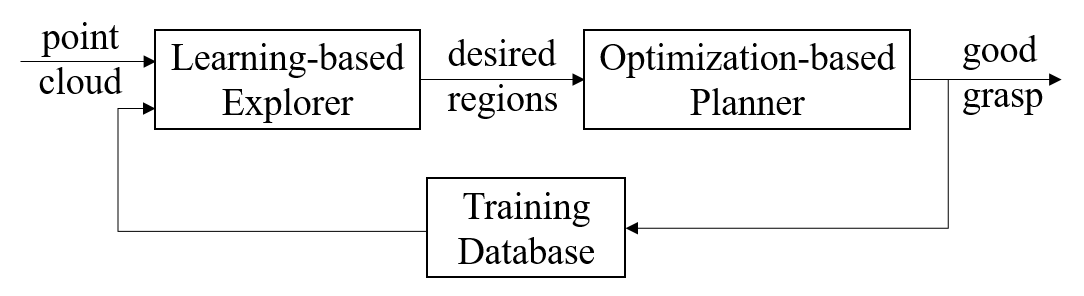}}
		\caption{Block diagram of the overall learning framework with RCNN.}
		\label{fig:learning_framework}
	\end{center}
\end{figure}

Figure~\ref{fig:learning_framework} shows the overall learning framework. 
The proposed learning framework decouples the end-to-end learning into a low-level optimization-based planner and a high-level learning-based explorer. The optimization-based planner searches for optimal gripper pose and finger configuration with the iterative surface fitting~(ISF) we proposed in~\cite{fan2018grasp}. The learning-based explorer detects the desired region to explore from the previous grasping experiences with the regions with R-CNN. Compared with end-to-end learning methods, the optimization-based planner is more precise and reliable when grasping on various unknown objects, and the learning-based explorer is more efficient with much less training data and lower learning dimension. 

\subsection{Optimization-based Planner}
\label{sec:optimization_based_planner}
\begin{figure}[t]
	\begin{center}
		{\includegraphics[width =3.4in]{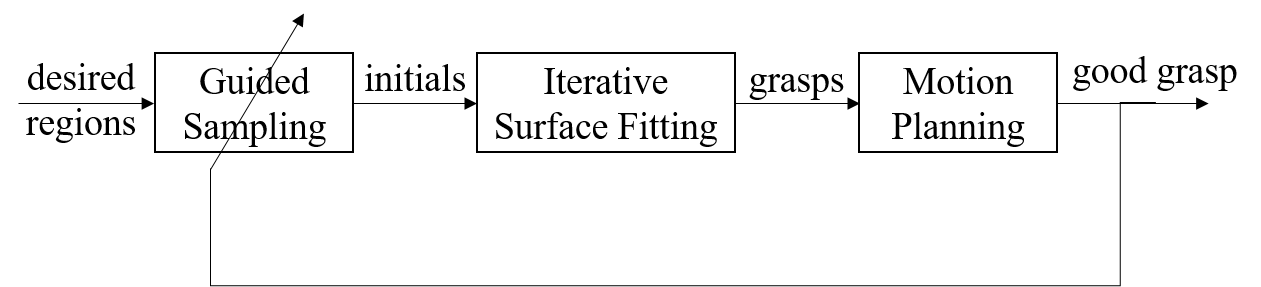}}
		\caption{Block diagram of the low-level optimization-based planner. }
		\label{fig:framework}
	\end{center}
\end{figure}

The optimization-based planner searches for the optimal gripper pose and finger configuration within some particular regions on the object surface. The block diagram of the optimization-based planner is shown in Fig.~\ref{fig:framework}.   It contains an iterative surface fitting (ISF) and a guided sampling. ISF is to register multiple fingertip surfaces to the target workpiece considering the allowable motions of fingers, and the guided sampling is to guide the ISF to search in different portions within the region to reduce the effect of the local optima.

\subsubsection{Iterative Surface Fitting}

\begin{figure}[bt]
	\begin{center}
	    {\includegraphics[width =1\linewidth]{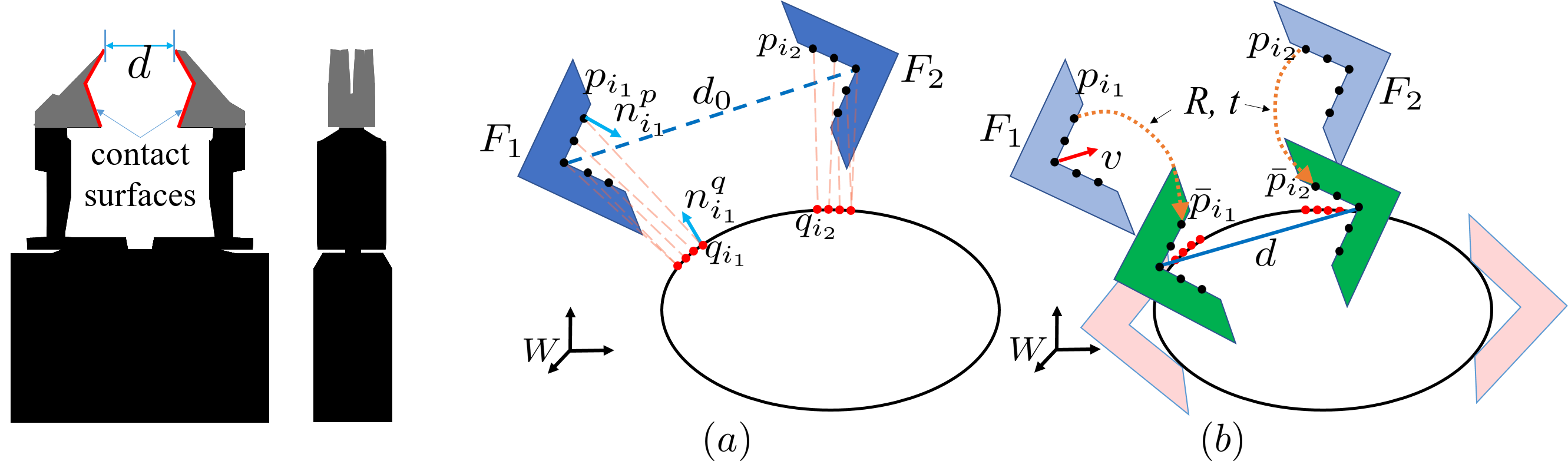}}
		\caption{(Left) Customized gripper used in the paper, and (Right) illustration of the iterative surface fitting (ISF). (a) The correspondence matching, where the corresponding point $q_{i_j}$ on the object surface is found by searching with the nearest neighbor and removing outliers and duplicates. (b) The surface fitting, in which the gripper transformation $(R,t)$ and finger displacement $\delta d$ are optimized. The green gripper is updated by the optimization result, and the pink one is the converged result after several iterations of ISF.}
		\label{fig:isf}
	\end{center}
\end{figure}

The algorithm is illustrated by a customized gripper with two fingers and one DOF, where the two fingers move in opposite directions to adjust the jaw width, as shown in Fig.~\ref{fig:isf} (Left). ISF iteratively executes two modules: the correspondence matching and the surface fitting. The correspondence matching contains the nearest neighbor search and outlier/duplicate filtering, as shown in Fig.~\ref{fig:isf} (Right)(a). 
The surface fitting searches over the gripper transformation $(R,t)$ and the jaw width $d = d_0 + \delta d$, as shown in Fig.~\ref{fig:isf} (Right)(b). More specifically, the surface fitting is to minimize the surface fitting error $E$: 
\begin{subequations}
	\label{eq3:overall}
	\begin{align}
	\min_{R, t, \delta d} &\  E(R,t,\delta d) \label{eq3:cost}\\
	s.t. \quad 
	& \delta d + d_0 \in [d_\text{min}, d_\text{max}] \label{eq3:surface_finger}
	\end{align}
\end{subequations}
Problem (\ref{eq3:overall}) is nonlinear due to the coupling between the $R$ and $\delta d$ and can be solved by an iterative palm-finger optimization (IPFO)~\cite{fan2018grasp}. 

The ISF algorithm is summarized in Alg.~\ref{alg:isf}. Inspired by~\cite{jost2003multi}, ISF is optimized hierarchically by searching with a multi-resolution pyramid. The inputs to ISF include the initial gripper state $R_c, t_c, d_0$, the surfaces $\partial \mathcal{O}$ and $\partial \mathcal{F}$. $L, I_0$ and $\epsilon_0$ denote the level number, maximum iteration and error bound for convergence, respectively.  The gripper surface $\partial F$ is first transformed to the specified initial state~(Line~\ref{isf:init}). In each level of pyramid, the $\partial \mathcal{F}$ is downsampled adaptively to $\mathcal{S}^f$ with different resolutions (Line~\ref{isf:down}). The while loop iteratively searches correspondence and solves for desired palm transformation and finger displacement. The while loop is terminated if IPFO gives a similar transformation in adjacent iterations.

\begin{algorithm}[tb]
	\caption{Iterative Surface Fitting (ISF)}\label{alg:isf}
	\begin{algorithmic}[1]
		\State \textbf{Input:} Initial state $R_c,t_c, d_0$, $\partial \mathcal{O}$, $\partial \mathcal{F}$, $L, I_0, \epsilon_0$ \label{isf:input}
		\State \textbf{Init:} $\partial \mathcal{F} = \mathcal{T}(\partial \mathcal{F}; R_c, t_c, d_0)$ \label{isf:init}
		\For {$l = L-1, \cdots, 0$} \label{isf:paraymid}
		\State $\mathcal{S}^f\leftarrow \texttt{downsample}(\partial \mathcal{F}, 2^l)$,  $I_l = I_0/2^l$,  $\epsilon_l= 2^l\epsilon_0$\label{isf:down}
		\State $\mathcal{S}_{0}^f \leftarrow \mathcal{S}^f$, $e_{s} \leftarrow \infty$, $\eta \leftarrow 0$
		\While {$\eta \notin [1 - \epsilon_l, 1 + \epsilon_l] $ and $it\texttt{++} \leq I_l$}
		\State $e_{s,p} \leftarrow e_s$
		\State $\mathcal{S}^o\leftarrow NN_{\partial O}(\mathcal{S}^f)$  \label{isf:nn}
		\State $\{\mathcal{S}^f, \mathcal{S}^o\} \leftarrow \texttt{filter}(\mathcal{S}^f, \mathcal{S}^o)$ \label{isf:filter}
		\State $\{R^*, t^*, \delta d^*, error\} \leftarrow \mathbf{IPFO}(\mathcal{S}^f, \mathcal{S}^o, d_0)$ \label{isf:IPFO}
		\State $\mathcal{S}^f \leftarrow \mathcal{T}(\mathcal{S}^f; R^*, t^*, \delta d^*)$ \label{isf:updateS} 
		\State $\partial \mathcal{F} \leftarrow \mathcal{T}(\partial \mathcal{F}; R^*, t^*, \delta d^*)$ \label{isf:update}
		\State $d_0 \leftarrow d_0 + \delta d^*$
		\State $e_s \leftarrow \|\mathcal{S}^f - \mathcal{S}_{0}^f\|, \ \eta \leftarrow e_s/e_{s,p}$ 
		\EndWhile
		\EndFor \label{isf:paraymid2} 
		\State \Return $\{ error, \partial \mathcal{F}\}$
	\end{algorithmic}
\end{algorithm}

\subsubsection{Baseline Initialization and Guided Sampling}
\label{sec:guided_sampling}
The initialization of ISF is important since ISF converges to local optima. Various initial points are desired for ISF to explore different portions of the region, so as to avoid getting trapped in bad local optima and achieve better collision-avoiding solutions. 
In the baseline initialization, the region is partitioned into $K$ clusters by k-means clustering. The center of each cluster is regarded as a candidate initial position of the gripper for ISF. The ISF with this baseline initialization is called baseline-ISF in the remaining of the paper. 

We build an empirical model to guide the sampling among $K$ candidates. Similar to the multi-armed bandit model, we record the fitting error, collision of ISF, and the reachability for each cluster and compute the average regret accordingly. The weight for ISF to be initialized in the $k$-th center is decreased if this center has larger regret value, and the cluster center with the minimum regret is chosen as the initial position of the gripper for the following ISF, while the initial orientation $R_c$ is randomly sampled.

\begin{figure}[t]
	\begin{center}
		{\includegraphics[width =1\linewidth]{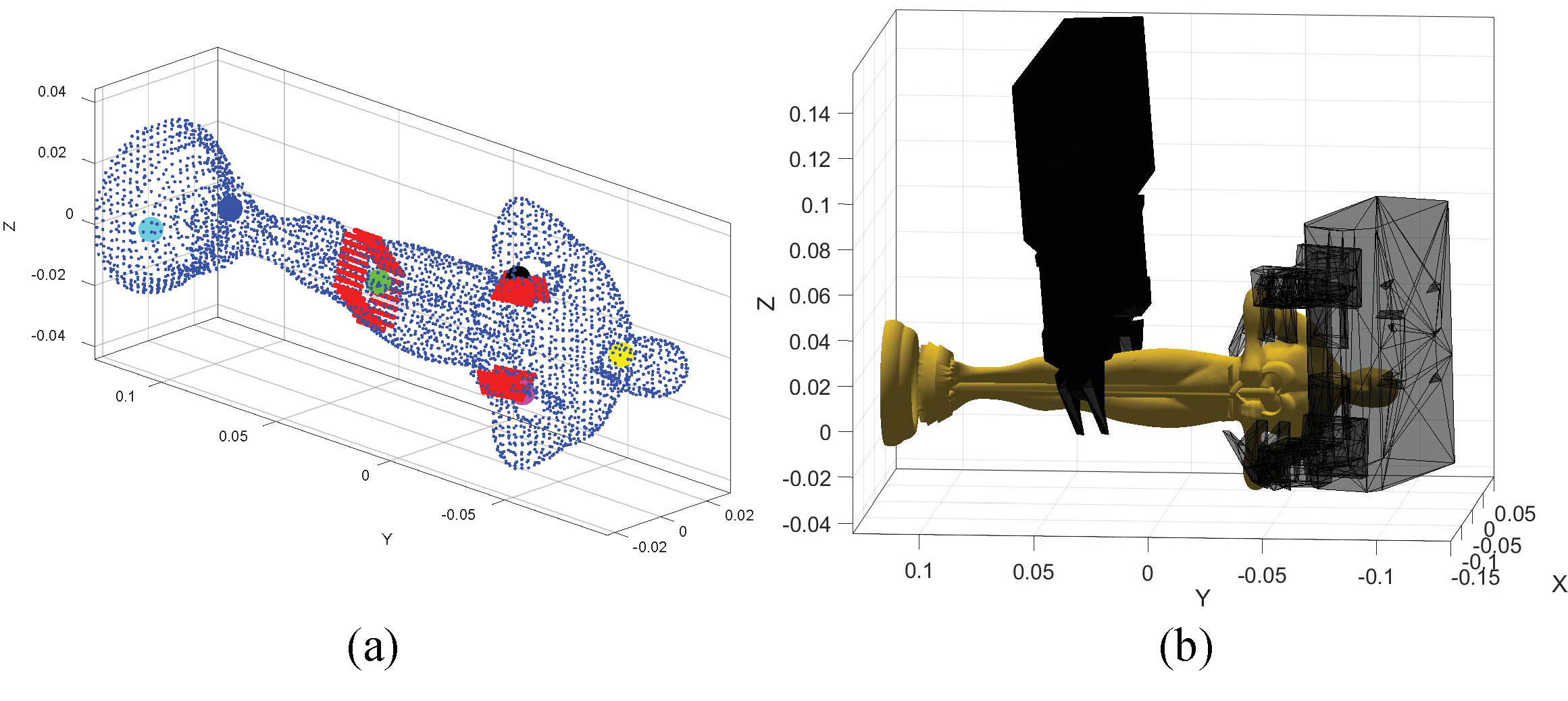}}
		\caption{ (a) Illustration of the grasp planning on an Oscar model, where the blue and red dots represent the object and the gripper surfaces, and the bold dots represent the centers of k-means. (b) The visualization of planned grasps, where the collided grasp is represented by the transparent one.}
		\label{fig:with_wo_collision}
	\end{center}
\end{figure}

Figure~\ref{fig:with_wo_collision} shows the baseline-ISF result on an Oscar model. The object and gripper surfaces are fed into the algorithm as shown by the blue and red dots in Fig.~\ref{fig:with_wo_collision}(a), after which the k-means clustering ran for centers of initialization, as shown by bold dots. 
Multiple grasps were generated (red patches in~Fig.~\ref{fig:with_wo_collision}(a)) and passed through the collision check function. The planned collided grasp is shown with transparency, as shown in  Fig.~\ref{fig:with_wo_collision}(b). The centers with small surface fitting error and no collision would have small regret values, thus will be sampled more frequently. 

The optimization-based planner (baseline-ISF) with the k-means clustering and guided sampling is able to search the optimal collision-free grasps efficiently in simple environment with small regions. However, the searching would be less efficient and sub-optimal in clutter environment where an excessively large number of clusters are required to be searched. Moreover, while the guided sampling distinguishes different centers of clusters by taking into account of the fitting error, collision and reachability, this allocation only explores around the predefined $K$ positions without considering other portions of the region. Finally, the guided sampling can only exploit the experience of the current grasp searching in the current environment, since the previous grasping is conducted in different environment with different distribution of centers and regrets.

\subsection{Learning-based Explorer}
\label{sec:Learning-based Explorer}
The optimization-based planner solved by baseline-ISF is inefficient and sub-optimal in clutter environment. Meanwhile, we found that human tends to decouple the process of choosing the desired grasp region from that of searching specific grasp poses inside that region to increase the efficiency of the grasping. With this observation, we design a learning-based explorer to initialize the baseline-ISF search. The explorer selects the regions with potential low regret based on the previous grasp experiences. We use R-CNN to learn a classifier in order to detect the desired regions for initialization.       
\subsubsection{R-CNN Pipeline} 
\begin{figure}[t]  
	\centering
    {\includegraphics[width =1\linewidth]{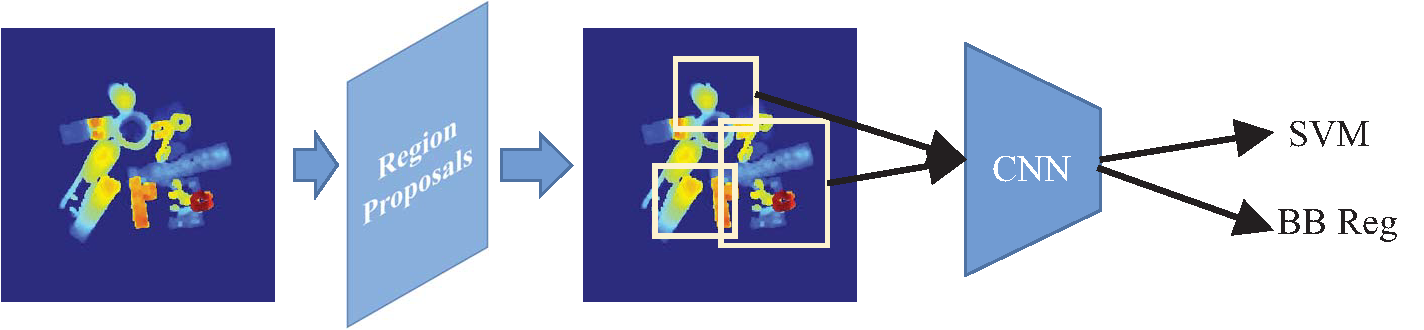}}
	\caption{Illustration of R-CNN pipeline. A region proposal block proposes regions and sends to CNN to extract features. SVMs and bounding box regression are applied to classify the proposed region and refine the bounding boxes, respectively. }
	\label{fig:rcnn}
\end{figure}

The pipeline of R-CNN is shown in Fig.~\ref{fig:rcnn}. R-CNN is first introduced in~\cite{girshick2014rich} for object detection. R-CNN contains a region proposal block to provide possible choices of regions.  The region proposal selects 2K regions with different sizes using the method such as selective search. The regions are resized and fed into a CNN for feature extraction. The CNN can be pre-trained by AlexNet~\cite{Krizhevsky2012ImageNet} or VGG-16/19~\cite{simonyan2014very}. The outputs of CNN are used to represent the features of the region proposals. SVMs are applied to classify the regions and bounding box regression is applied to further correct the positions of the bounding boxes.  

\subsubsection{R-CNN Training}
\begin{figure}[t]  
	\centering
	{\includegraphics[width =1\linewidth]{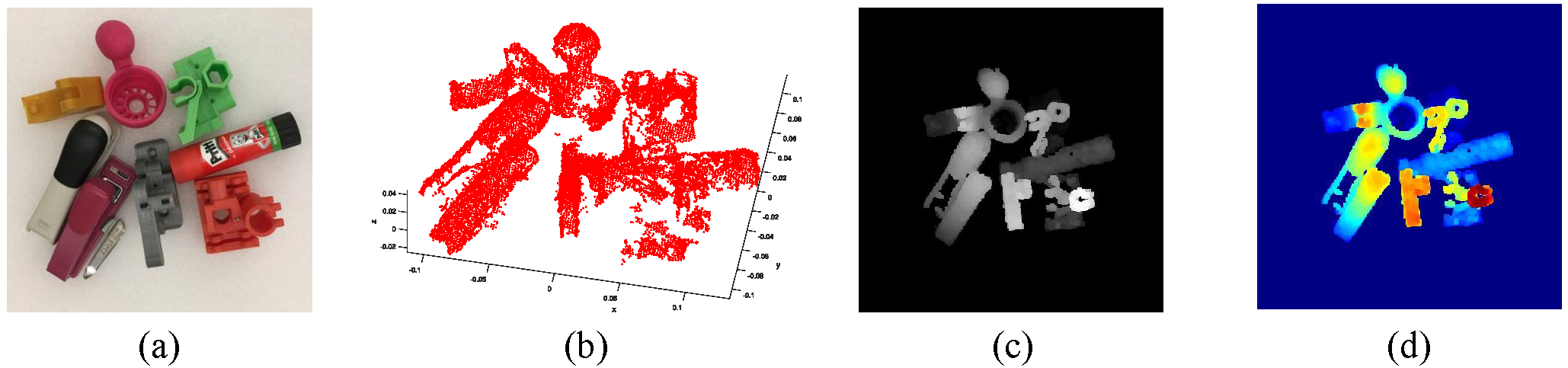}}
	\caption{The depth rendering. (a) Original scene. (b) Point cloud observed by two stereo cameras. (c) Rendered depth images. The closer of the points to the camera, the more white they are in the depth image. (d) The image is further rendered into jet colormap. }
	\label{fig:depthImgRendering}
\end{figure}

\begin{figure}[t]  
	\centering
	{\includegraphics[width =1\linewidth]{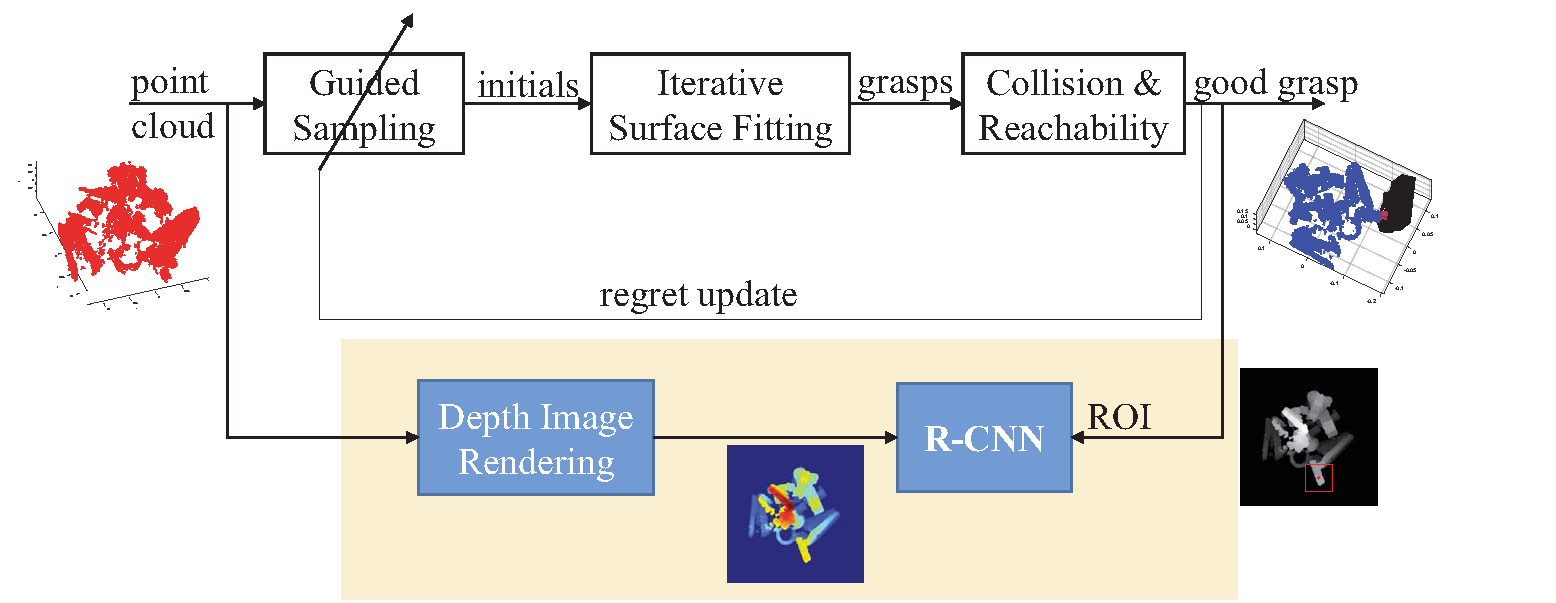}}
	\caption{Illustration of the training framework. The framework contains the baseline-ISF and the R-CNN training. }
	\label{fig:training_framework}
\end{figure}

\begin{figure}[t]  
	\centering
	{\includegraphics[width =1\linewidth]{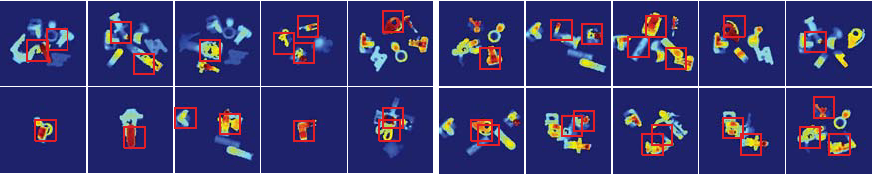}}
	\caption{The data used to fine-tune the R-CNN for good region detection.}
	\label{fig:training_data}
\end{figure}

We use a grasping pool with 25 objects of different categories. In industrial bin picking applications, this number is usually sufficiently large. 
We randomly choose some of the objects and place them in the workspace. The scene is observed by two stereo cameras and the collected point cloud is rendered as depth images, as shown in Fig.~\ref{fig:depthImgRendering}.  
R-CNN takes the rendered depth images as inputs, and produces regions of interest (ROI) to initialize the ISF searching. The ROI used for training is generated based on the optimal grasps found from baseline-ISF. The training process is illustrated by~Fig.~\ref{fig:training_framework}. 
The R-CNN in this paper is pre-trained by AlexNet and is fine-tuned by the data we collected. In this stage, we use 250 data pairs with data augmentation, producing 2000 data pairs to fine tune the network. Some of the data pairs are illustrated in Fig.~\ref{fig:training_data}. 

\subsubsection{R-CNN Testing}
\begin{figure}[t]  
	\centering
	{\includegraphics[width =1\linewidth]{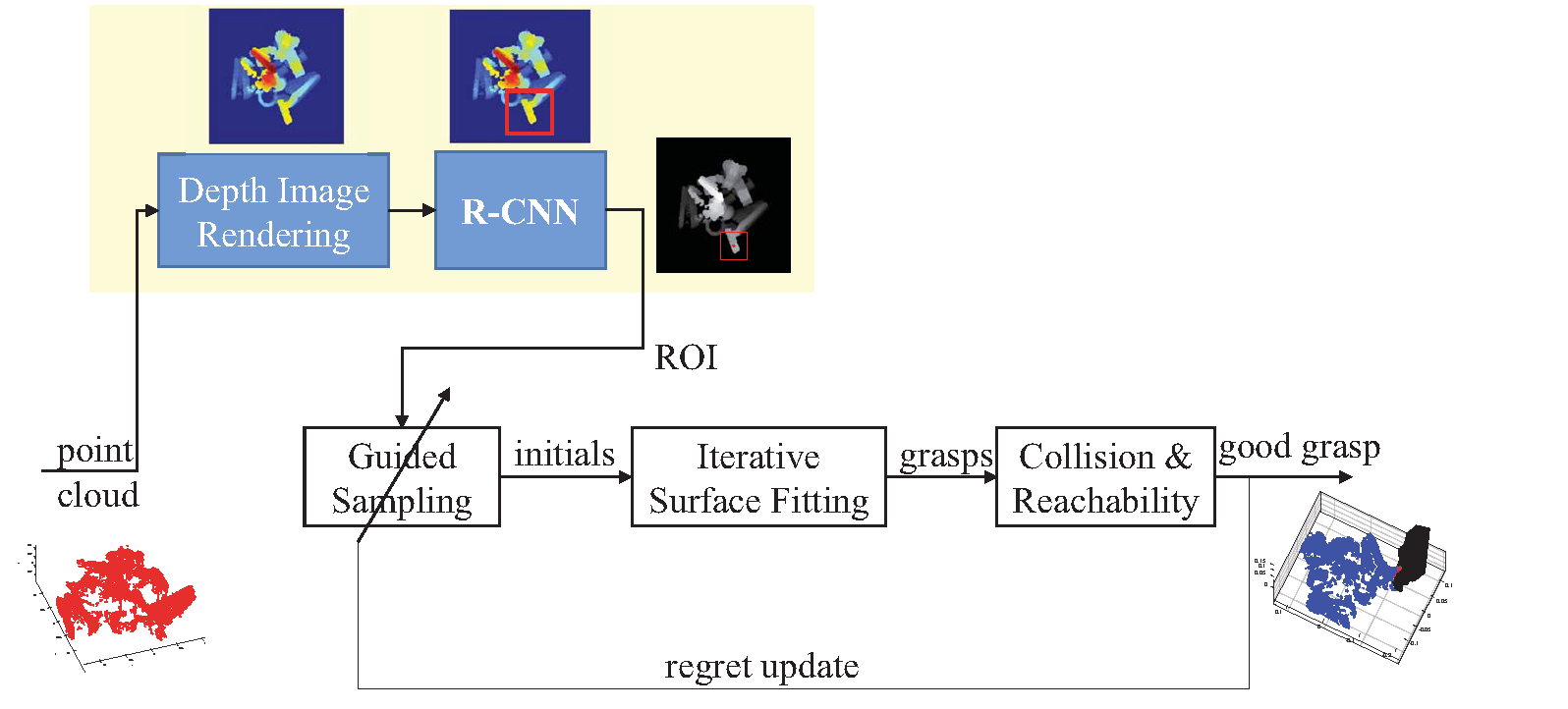}}
	\caption{The learning framework with RCNN-ISF implementation. The R-CNN detects desired low regret regions and ISF searches grasps on the selected regions.}
	\label{fig:test_framework}
\end{figure}

At the test time, the trained R-CNN is applied to generate the ROI on the colormap. The desired regions in the point cloud are then computed based on the box coordinates of ROI in the image plane as well as their depth values. 
The centers of these regions are regarded as the good initialization and ISF searches from these initial positions. 
The proposed learning ISF is called RCNN-ISF and the framework is illustrated in Fig.~\ref{fig:test_framework}. The proposed learning framework with RCNN-ISF not only searches grasps locally within a small region, but also learns the large-scale grasp exploration using R-CNN. Therefore, it tends to provide better initialization more efficiently than the baseline-ISF. 

Compared with end-to-end learning~\cite{mahler2016dex,levine2016learning}, the proposed RCNN-ISF method has the following advantages. First, the learning dimension of the RCNN-ISF is generally lower than that of the end-to-end learning. More specifically, RCNN-ISF searches ROI in the two-dimensional image plane, while the end-to-end learning searches over higher dimension depending on the grasps and grippers. For example, a grasp planning for a eight-DOF hand with three fingers has 32 dimensions~\cite{fan2018real}. 
Therefore, the end-to-end learning requires much more data than the proposed method. 
Secondly, ISF searches for optimal grasps based on object-specific features that are not shared cross objects, instead of learning the behavior end-to-end from millions of data. Consequently, ISF tends to generate more precise and robust grasps. 
Moreover, RCNN-ISF is able to produce versatile grasp as the experiment shows. On the contrary, the learned networks in~\cite{mahler2016dex, levine2016learning} produce planar grasps with simple parallel grippers.

\section{Experimental Results}
\label{sec:experimental_results}
This section presents the experimental results for baseline-ISF and RCNN-ISF to verify the effectiveness of the learning framework. The experimental videos are available at~\cite{website}. The host computer we used was a desktop with 32GB RAM and 4.0GHz CPU. All the computations were conducted by Matlab. 
We used a SMC LEHF20K2-48-R36N3D parallel jaw gripper with the specialized curved fingers as shown in Fig.~\ref{fig:isf}~(Left). The desired contact surfaces are marked by red. The gripper was equipped on a FANUC LRMate 200iD/7L industrial manipulator. Two IDS Ensenso N35 stereo camera sets were used to capture the point cloud of the object. The point cloud was smoothed and the normal vectors were calculated. Despite the preprocessing, the point cloud produced by Ensenso cameras was not able to reflect the object precisely due to occlusion and noise. 

\subsection{Parameter Lists}
The gripper width was constrained by $\left[d_\text{min},d_\text{max} \right] = \left[1,3\right]$ cm. The initial gripper width was set as $d_0$ = 2 cm. The pyramid level $L$, the maximum iteration $I_0$ and the tolerance $\epsilon_0$ were set as $4$, $200$ and $0.008$ respectively in Alg.~\ref{alg:isf}. The k-means center number $K = 6$. 

\subsection{Baseline-ISF Experiment}
\label{sec:baseline-isf}
This section presents the experimental results of the optimization-based planner with baseline-ISF. The baseline-ISF algorithm considers the surface fitting around different initial positions specified by k-means. The guided sampling enables more effective exploration based on the grasp experience of the current environment. 

\begin{figure}[t]  
	\centering
	{\includegraphics[width =1\linewidth]{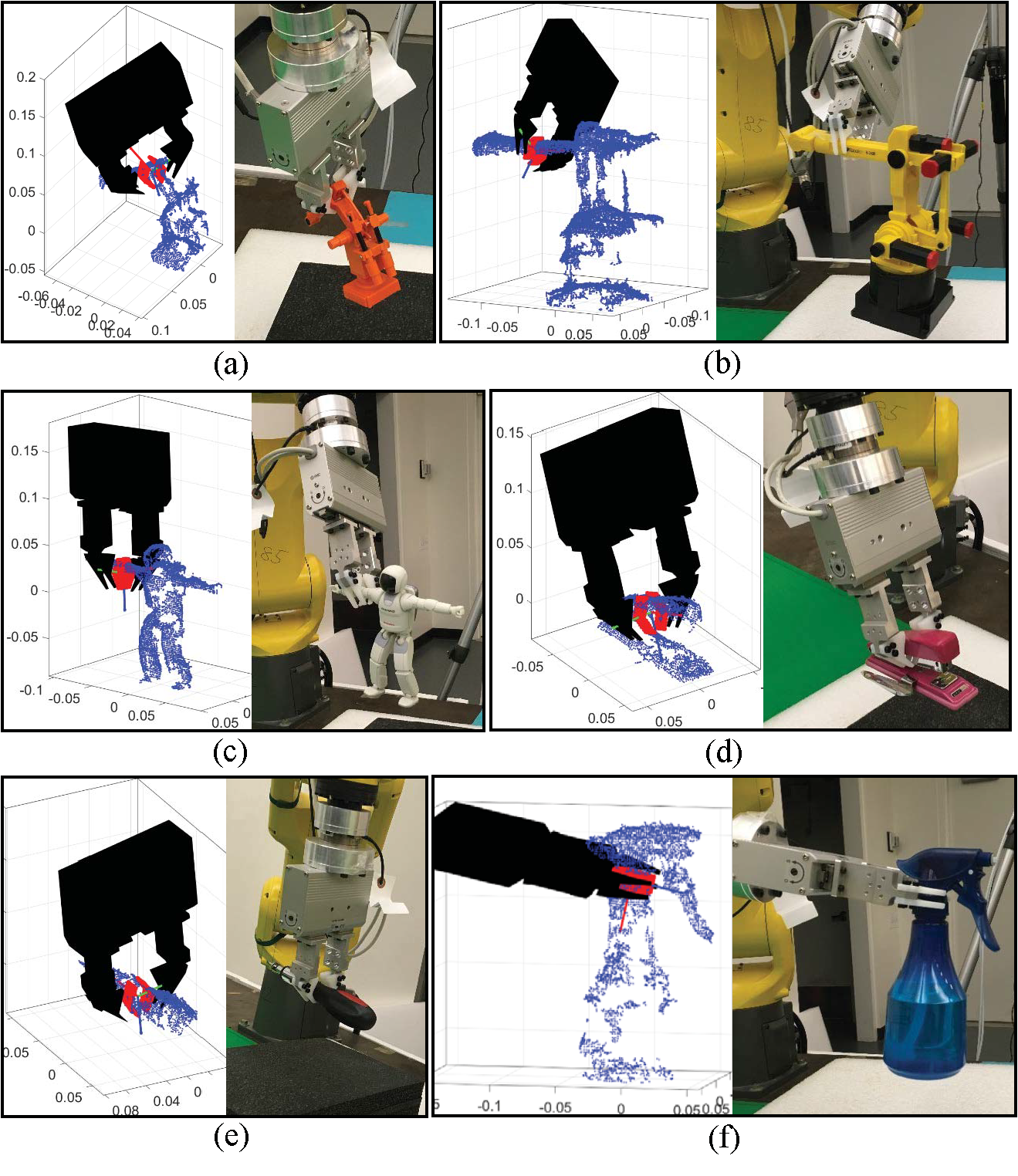}}
	\caption{The results of the grasp planning experiment on six objects.}
	\label{fig:graspExp1}
\end{figure}

Figure~\ref{fig:graspExp1} shows the grasp planning results on six different objects. The left and right sides of each subfigure show the grasp optimized by ISF and the result in lifting the object by 10 cm, respectively. Baseline-ISF was able to find a grasp that matched the fingertips to the object in spite of the complicated shapes of the surface. The algorithm was able to handle the missing points properly. For example, the gripper fingertip was able to locate to the bottom of the link (Fig.~\ref{fig:graspExp1}(b)), despite the points were missing since both cameras were looking from the top. 
The geometry similarity between the gripper and the object is beneficial for the gripper to resist the effect of calibration uncertainty and external disturbance, and increase the stability and the payload, as it increases the potential contact surfaces between the gripper and the object. 
In average, it takes less than 1.5 secs to find 10 collision-free optimal grasps. 

\subsection{RCNN-ISF Experiment}
\label{res:rcnn_exp}
This section presents the experimental results of the proposed learning framework with RCNN-ISF. The experimental videos are available at~\cite{website}. The learning framework includes the low-level optimization-based planner with baseline-ISF, and the high-level learning-based explorer with R-CNN for grasp exploration in heavy clutter environment. The training process of R-CNN is shown in Fig.~\ref{fig:training_data}. Same experimental setup was used as Section~\ref{sec:baseline-isf}. 

\begin{figure}[t] 
	\centering
    {\includegraphics[width =1\linewidth]{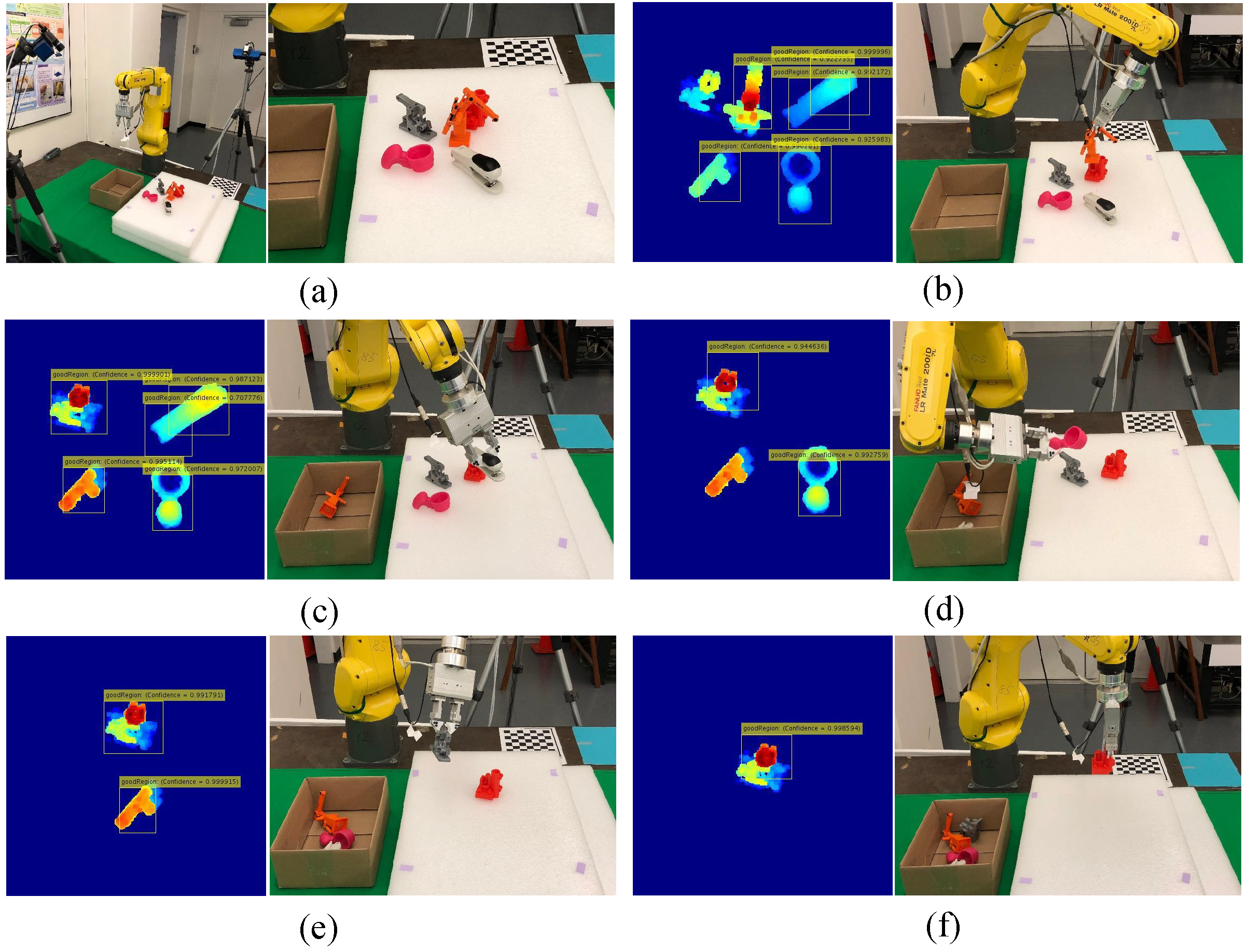}}
	\caption{Grasp planning experiment in a clutter environment with the proposed learning framework. (a) The initial object clutter. (b)-(f) The consecutive grasps in the task.}
	\label{fig:graspExp2}
\end{figure}

Figure~\ref{fig:graspExp2} illustrates the performance of the RCNN-ISF in light clutter environment. In this environment, RCNN-ISF achieved comparable performance with the baseline-ISF. 
The initial configuration of the object set is shown in Fig.~\ref{fig:graspExp2}(a). Figure~\ref{fig:graspExp2}(b)-(f) show the consecutive grasps in the task. The left side of each subfigure is the depth image and the R-CNN output of the confidence map for the desired regions, after which the baseline-ISF was performed on the chosen regions, and the best grasp was executed by the FANUC industrial manipulator, as shown in the right side of each subfigure. 
Even though the surface composed by the cluttered objects became more complicated than a single object, RCNN-ISF was able to successfully detect the desired regions for grasping and search on the selected regions for the optimal collision-free grasp. 

\begin{figure}
    \centering
    \includegraphics[width = 1\linewidth]{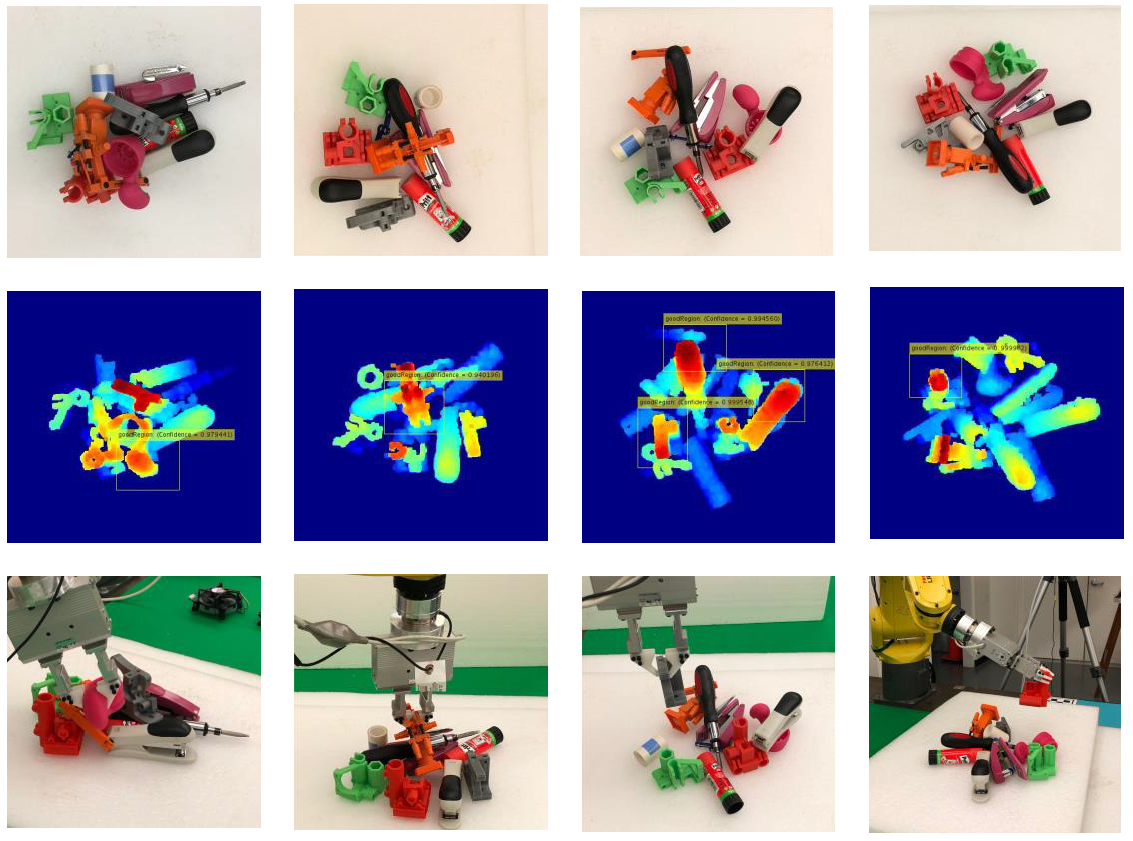}
    \caption{Grasp planning results on four different heavy clutter environments by RCNN-ISF. (Up) Clutter environments. (Middle) Selected regions and their confidence values by R-CNN to initialize ISF searching. (Bottom) Execution of the grasps planned by ISF. }
    \label{fig:GraspExp_RCNN}
\end{figure}

Figure~\ref{fig:GraspExp_RCNN} shows the grasp planning results on heavy clutter environments by RCNN-ISF. The first row shows different clutter environments. R-CNN took rendered depth images as inputs and generated the desired regions to initialize ISF searching, as shown in Fig.~\ref{fig:GraspExp_RCNN} (Middle). The optimization-based planner produced optimal grasp pose by minimizing the fitting error and checking the collision/feasibility. The optimal grasps were executed by the FANUC manipulator, as shown in Fig.~\ref{fig:GraspExp_RCNN} (Bottom). 

The comparison between the baseline-ISF and RCNN-ISF is shown in Table~\ref{tab:comparison} for 4 heavy clutter environments in Fig.~\ref{fig:GraspExp_RCNN}.
The searching on heavy clutter environments are more difficult due to the collision with environments and the excessive search space for guided sampling. Consequently, the initialization becomes more important for efficient exploration. The baseline-ISF requires to initialize and execute ISF multiple times in order to explore broader regions. On the contrary, RCNN-ISF employs the previous experience for initialization and generates only a few low-regret regions to start ISF searching. Therefore, RCNN-ISF is able to perform grasp planning more efficiently. 
In heavy clutter environment, the baseline-ISF spent 17.23 secs to find 1 optimal grasp, while the proposed RCNN-ISF spent 1.52 secs to find 7 optimal grasps. 
\begin{table}[t]
\centering
\caption{Comparison of Baseline-ISF and RCNN-ISF}
\label{tab:comparison}
\begin{tabular}{r|cc}
Methods & Baseline-ISF & RCNN-ISF \\
\hline 
Search Time (s)        & 17.23  & 1.52  \\
\hline 
Found Grasps\#     & 1 & 7 \\
\hline
\end{tabular}
\end{table}

\section{Conclusions and Future Works} 
\label{sec:conclusions}
This paper proposed a learning framework to plan robust grasps for customized grippers. The learning framework includes a low-level optimization-based planner and a high-level learning-based explorer.   
The optimization-based planner used an iterative surface fitting (ISF) with guided sampling to search for optimal grasps by minimizing the surface fitting error. The performance of this low-level planner was locally effective and was sensitive to initialization. Therefore, the learning-based explorer was introduced with a region-based convolutional neural network (R-CNN) to search for desired low-regret regions to initialize ISF search. A series of experiments on robotic bin picking were performed to evaluate the proposed method. Experimental results show that the proposed learning framework with RCNN-ISF achieved a more efficient planning on heavy clutter environment, by significantly decreasing the average searching time from 17.23 secs to 1.52 secs.  

There are several directions that we want to explore in the future. First, we would like to apply the algorithm to the customized grippers with more complicated shapes. In addition, we would like to employ the domain randomization~\cite{tobin2017domain} and train the learning-based explorer sufficiently and efficiently.

\addtolength{\textheight}{-1cm}   


\bibliographystyle{IEEEtran}
\bibliography{references}

\end{document}